\documentclass[letterpaper]{article} 
\usepackage{aaai2026}  
\usepackage{times}  
\usepackage{helvet}  
\usepackage{courier}  
\usepackage[hyphens]{url}  
\usepackage{graphicx} 
\usepackage{natbib}  
\usepackage{caption} 
\nocopyright
\usepackage{algorithm}
\usepackage{algorithmic}
\usepackage{subcaption}
\usepackage{amsmath}
\usepackage{newfloat}
\usepackage{listings}
\DeclareCaptionStyle{ruled}{labelfont=normalfont,labelsep=colon,strut=off} 
\floatstyle{ruled}
\newfloat{listing}{tb}{lst}{}
\floatname{listing}{Listing}
\title{Mapping Clinical Doubt: Locating Linguistic Uncertainty in LLMs}
\author{
    Srivarshinee Sridhar\textsuperscript{\rm 1}\equalcontrib,
    Raghav Kaushik Ravi\textsuperscript{\rm 1}\equalcontrib,
    Kripabandhu Ghosh\textsuperscript{\rm 2}\\
}
\affiliations{
    \textsuperscript{\rm 1} Vellore Institute of Technology, Chennai\\
    \textsuperscript{\rm 2} IISER Kolkata

}

\usepackage{bibentry}

\begin{document}

\maketitle

\begin{abstract}
Large Language Models (LLMs) are increasingly used in clinical settings, where sensitivity to linguistic uncertainty can influence diagnostic interpretation and decision-making. Yet little is known about where such epistemic cues are internally represented within these models. Distinct from uncertainty quantification, which measures output confidence, this work examines input-side representational sensitivity to linguistic uncertainty in medical text. We curate a contrastive dataset of clinical statements varying in epistemic modality (e.g., “is consistent with” vs. “may be consistent with”) and propose Model Sensitivity to Uncertainty (MSU), a layerwise probing metric that quantifies activation-level shifts induced by uncertainty cues. Our results show that LLMs exhibit structured, depth-dependent sensitivity to clinical uncertainty, suggesting that epistemic information is progressively encoded in deeper layers. These findings reveal how linguistic uncertainty is internally represented in LLMs, offering insight into their interpretability and epistemic reliability. 
\begin{links}
    \link{Code}{https://github.com/varshin5699/Mapping-Clinical-Doubt--Locating-Linguistic-Uncertainty-in-LLMs}
\end{links}
\end{abstract}


\section{Introduction}
\begin{figure}
    \centering
\includegraphics[width = 1\linewidth]{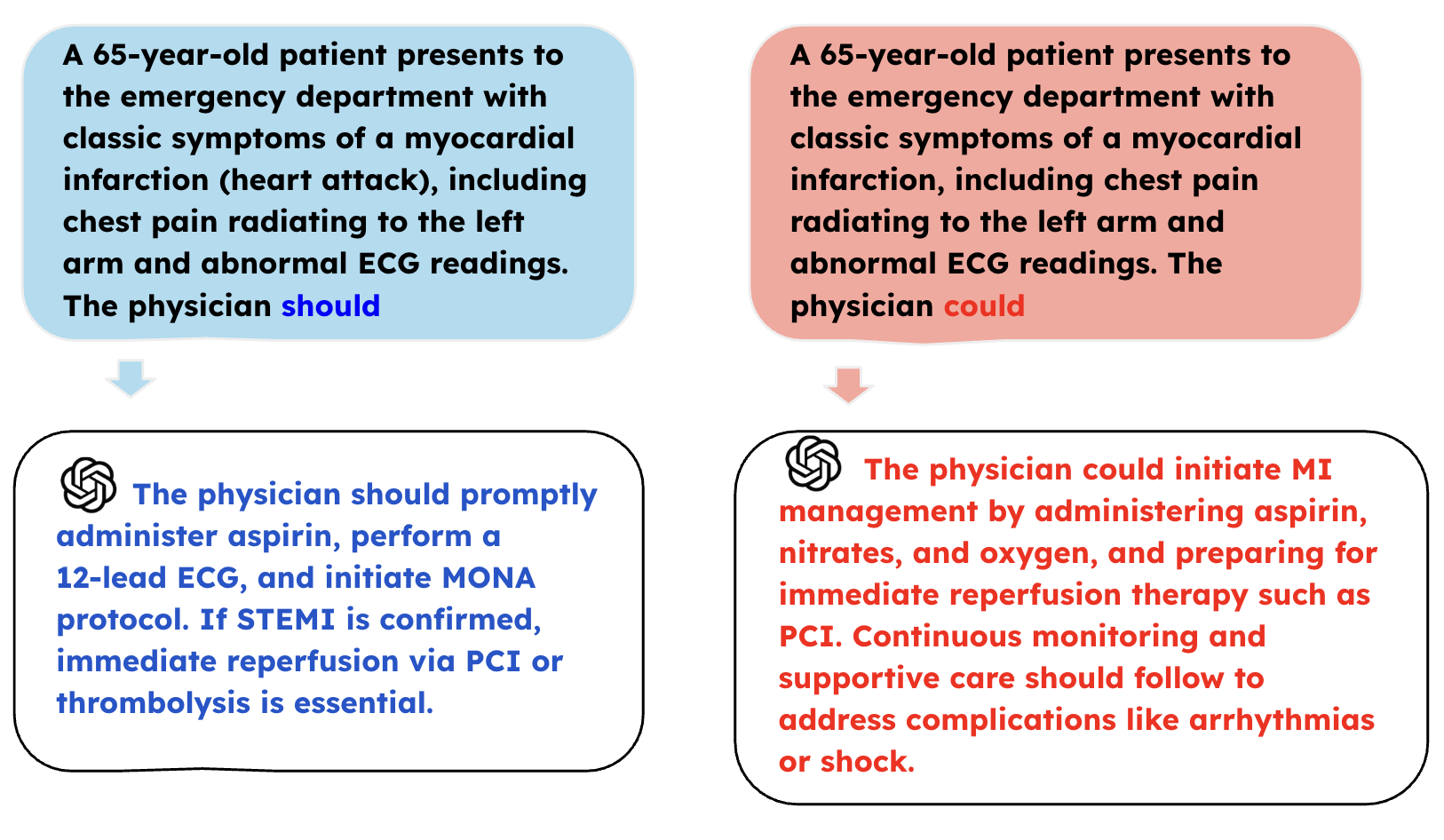}
    \caption{Although the prompt pairs differ only in epistemic modality (\textit{should vs could}), the responses vary: those prompted with \textit{could} tend to offer a broader range of medical possibilities and are more open-ended compared to those ending with \textit{should}. This could imply how the model interprets linguistic uncertainty.}
    \label{fig:epistemic-modality-example}
\end{figure}

Large Language Models (LLMs) are increasingly deployed in high-stakes domains such as law, medicine, and public policy settings that demand not only accurate outputs but also the ability to reason responsibly under uncertainty. While previous research has emphasized calibrating model confidence and evaluating output probabilities, comparatively little is known about how LLMs internally represent input-side uncertainty particularly linguistic uncertainty expressed through epistemic modality (e.g., might, could, probably). Figure~\ref{fig:epistemic-modality-example} shows that even minimal shifts in modality, such as replacing `should' with `could', lead to consistent differences in model generations, despite all other input remaining constant. These variations are not artifacts of sampling noise; rather, they indicate a systematic and grounded sensitivity to uncertainty cues. This raises important questions: How are such epistemic signals encoded across the layers of a model? Are these representations stable across model variants? Understanding the internal treatment of linguistic uncertainty is critical for building trustworthy, transparent systems, especially in applications where appropriately responding to hedged or speculative language can directly affect outcomes and user trust.

\subsection{Uncertainty in LLM Outputs}
Much of the recent work on LLM uncertainty has focused on model outputs, through calibration techniques \cite{desai-durrett-2020-calibration}, truthfulness under uncertainty \cite{lin2022teaching}, or confidence alignment with ground-truth \cite{ghafouri2024epistemic}. While these approaches provide useful diagnostics on model predictions, they do not investigate how input uncertainty is internally represented or whether models distinguish between certain and uncertain prompts at a representational level.

\subsection{Epistemics and Modal Reasoning}
Several recent studies have examined the role of modal verbs in model reasoning. \cite{holliday2024modal}) show that LLMs often struggle with logical tasks involving modal operators, suggesting a lack of systematic reasoning with modality. 


\cite{zhou2023uncertainty} analyze use of epistemic markers in LLM-generated text, showing large effects on accuracy depending on whether uncertainty or certainty markers are used, although they do not study how these markers are internally represented in neural activations. Similarly, \cite{lee2025llmjudges} show that LLM-based evaluators are systematically biased against responses containing expressions of uncertainty.

Our work complements these efforts by offering a mechanistic perspective on how epistemic modality is encoded inside models, using probing over activation spaces. In contrast to prior work focusing on usage or output alignment, we provide empirical evidence of internal sensitivity to epistemic variation.

\subsection{Probing Internal Representations}
A growing body of mechanistic interpretability research seeks to understand LLM internals by intervening in the \textit{activation space}. One core technique is activation patching, also known as causal mediation or causal tracing, which substitutes hidden activations from a clean forward pass into a corrupted run, thereby identifying components causally responsible for specific behaviors \citep{NEURIPS2020_92650b2e}. Building on this, path patching refines the approach by restricting interventions to specific computational paths, enabling finer-grained localization of functional subcircuits \citep{goldowskydill2023localizingmodelbehaviorpath}. Complementary methods include causal scrubbing \cite{chan2022causal}, which tests whether abstract circuits maintain function across structural perturbations, and automated circuit discovery frameworks such as ACDC \citep{conmy2023automated}.

\begin{figure}
    \centering
    \includegraphics[width=1\linewidth]{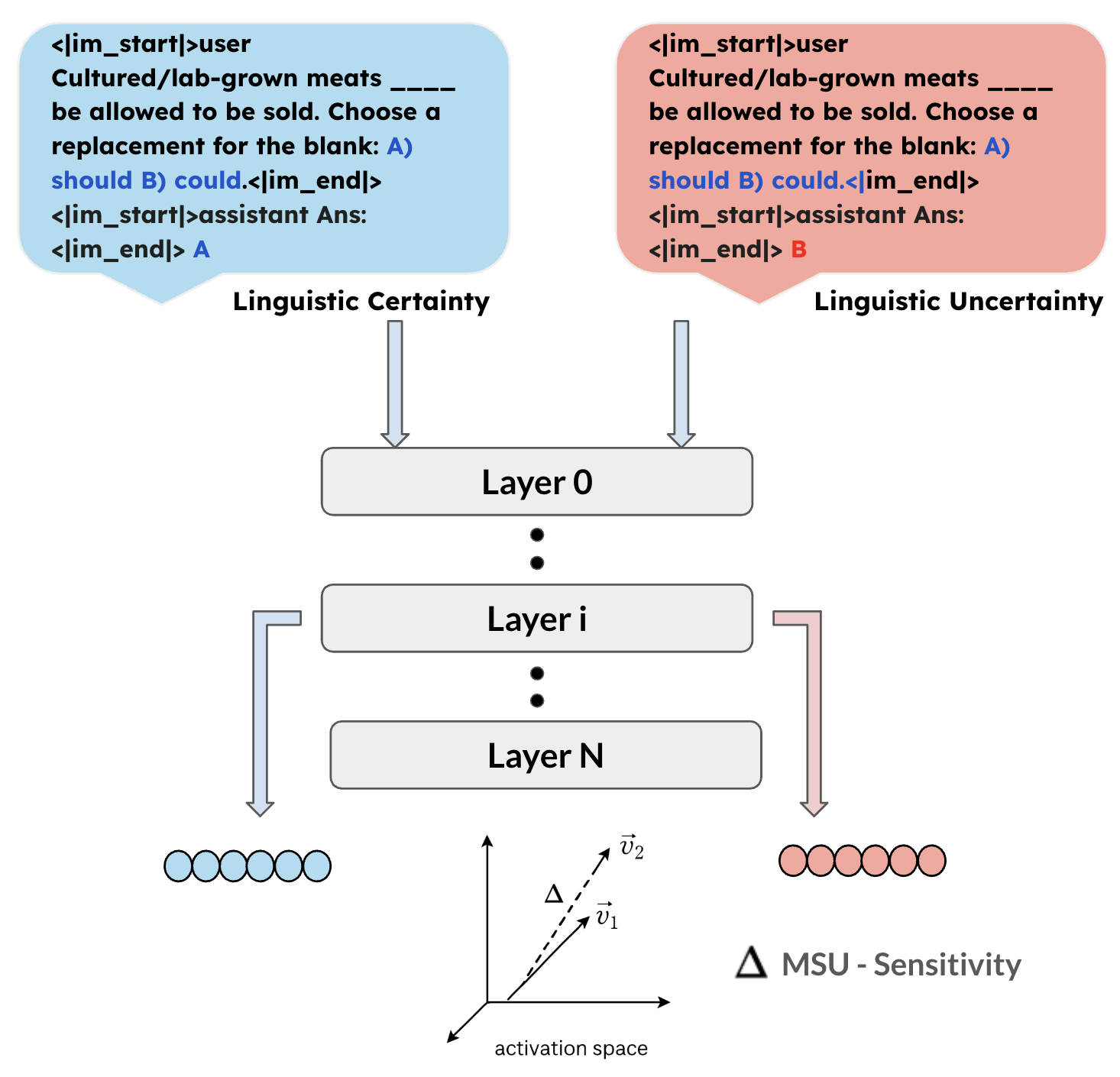}
    \caption{Paired inputs with linguistic certainty and uncertainty are passed through the model. Layerwise activation differences ($\Delta$) are used to compute MSU, capturing the model's sensitivity to uncertainty across layers.}
    \label{fig:hero-figure}
\end{figure}

In this work, we pose the question: \textit{Are large language models sensitive to epistemic modality in their input?} This question is investigated by contrasting representations elicited by semantically similar prompts that differ only in the degree of certainty they convey. As depicted in Figure \ref{fig:hero-figure}, we curate a multiple-choice dataset of 3,114 sentence pairs that differ in their expression of certainty and examine how these variations influence the model’s internal activation space.

To quantify such effects, we introduce a novel metric, \textbf{Model Sensitivity to Uncertainty (MSU)}, which captures the representational shift induced by epistemic cues at each layer of the model. Through this lens, we assess whether and where in the model epistemic modality is internally encoded.
Our key contributions are as follows:
\begin{itemize}
    \item We propose a probing framework to assess whether LLMs encode epistemic modality through changes in their activation space.
    \item We introduce \textbf{MSU}, a layerwise metric for quantifying model sensitivity to linguistic uncertainty.
    \item We release the dataset of 3,114 sentence pairs depicting linguistic certainty and uncertainty.
\end{itemize}

\section{Dataset}
Unlike contrastive datasets used in prior work \cite{rimsky-etal-2024-steering}, data was not constructed to be semantically opposite or adversarial. Instead, they vary along a fine-grained linguistic axis of uncertainty, making them well-suited for probing representational sensitivity to uncertainty.

The dataset used in this work is derived from claims in the \texttt{Anthropic/Persuasion} \cite{durmus2024persuasion} corpus. Sentences containing modal verbs such as ``should'' and ``must'' were identified and programmatically masked  using the \texttt{pandas} \cite{ mckinney-proc-scipy-2010}, \texttt{NumPy} \cite{harris2020array}, and \texttt{NLTK} \cite{bird-loper-2004-nltk} libraries. These masked positions were then filled with controlled multiple-choice options representing either certain or uncertain linguistic modality (e.g., ``should'' vs. ``could'').

For example, consider a sample from the \texttt{Anthropic/Persuasion} dataset \cite{durmus2024persuasion}:
\newline
\noindent
\fbox{%
\parbox{\linewidth}{%
\textbf{Example Prompt}

\textit{Original: ``Governments and technology companies must do more to protect online privacy and security.''}

\medskip
\textbf{Modified prompt:}

\begin{small}
\texttt{
<|im\_start|>user\\
Governments and technology companies\\
\texttt{[MASK]} do more to protect online\\
privacy and security.\\
Choose a replacement for the MASK.\\
A) Must \;\; B) Might\\
<|im\_end|>
}
\end{small}
}%
}
\\

One of the pair would be appended with A, and the other with B, as shown in Figure \ref{fig:hero-figure} to form the linguistic certain-uncertain pair. 
The changes made to the dataset ensure the following:

\noindent {\bf Controlled Variation:} The only systematic difference across each pair is the use of uncertainty markers (e.g., ``might'', ``possibly'', ``maybe''), ensuring minimal lexical confounds.

\noindent {\bf Semantic Stability:} As the core semantics is preserved, any variation in activation vectors can be more confidently attributed to modality, not content drift. 
For examples of these some of these changes, refer Appendix \ref{sec: dataset-examples}.

Each sentence was paired with two variants, one expressing certainty and one expressing uncertainty by appending the starting letter of the option at the end of the instruction, resulting in 3,114 samples per condition, and a total of 6,228 examples.

\section{Evaluation Setup}

We conduct our analysis on three small-scale language models: Qwen2.5-0.5B-Instruct \cite{qwen2.5}, Qwen1.5-0.5B-Chat \cite{qwen}, and LLaMA-3.2-1B-Instruct\cite{grattafiori2024llama3}. 

\subsection{Model Specifications}
\label{sec:model-specs}
These models were selected to cover a range of instruction-tuned and chat-oriented variants with varying parameter counts while maintaining manageable computational overhead. All models provide access to internal activations, which is crucial for our layer-wise representation analysis.
ageable computational overhead.
Table~\ref{tab:model-info} details the model sizes and number of layers for the  language models used in our experiments. All internal activations were accessed using the \texttt{TransformerLens} \cite{nanda2022transformerlens} library.

\begin{table}[ht]
\centering
\begin{tabular}{|l|c|c|}
\hline
\textbf{Model} & \textbf{Parameters} & \textbf{Layers} \\
\hline
Qwen2.5-0.5B-Instruct & 391M & 24 \\
Qwen1.5-0.5B-Chat & 308M & 24 \\
Llama-3.2-1B-Instruct & 1.1B & 16 \\
\hline
\end{tabular}
\caption{Model sizes, number of layers, and sources for LLM variants used in our analysis.}
\label{tab:model-info}
\end{table}

\subsection*{Can Linguistic Uncertainty be probed in the Activation Space?}

Principal Component Analysis (PCA) is applied layer by layer to the model activation vectors to examine whether the uncertainty is encoded linearly separable. For each layer, activations of certain and uncertain examples are projected onto the two main components using Scikit-learn \citep{scikit-learn}, allowing visualization of potential clustering patterns. This analysis serves as a diagnostic to assess the validity of the data set to study linguistic uncertainty (refer to appendix \ref{sec: pca-analysis}).

\subsection*{Results}


\begin{figure*}[t] 
    \centering
    \includegraphics[width = 1\linewidth]{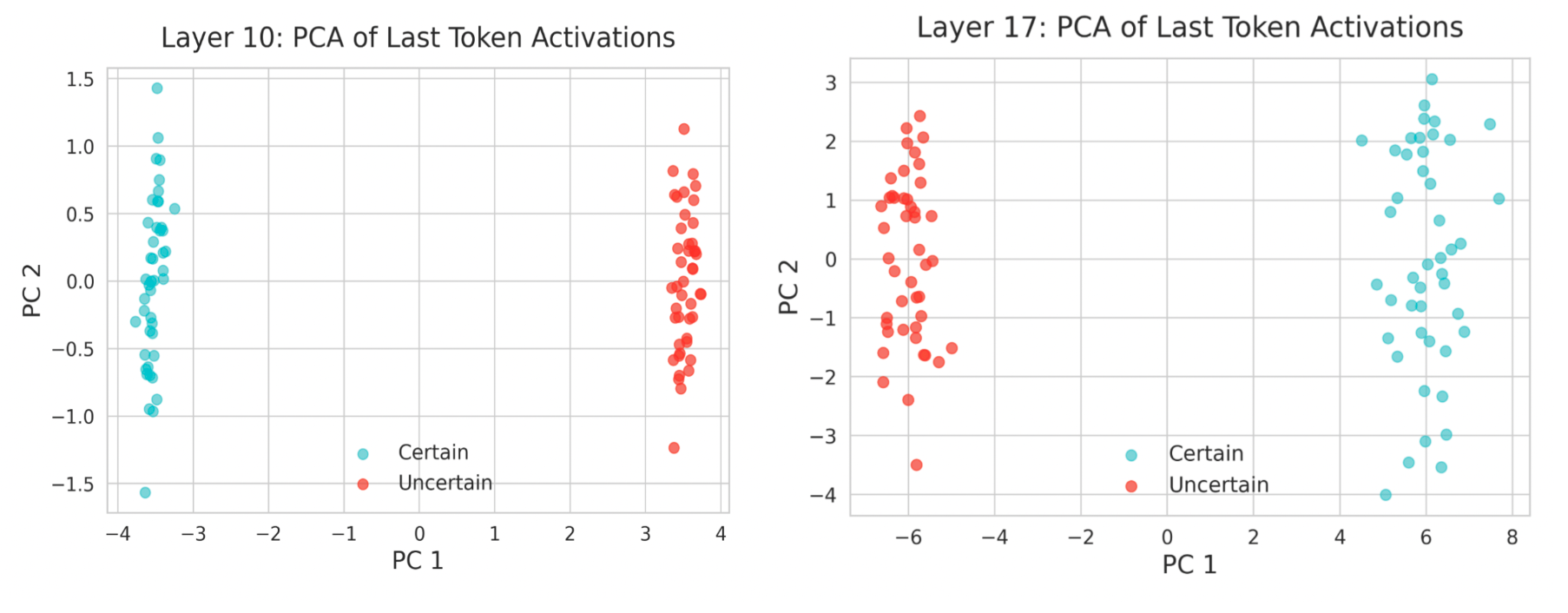}
    \caption{PCA plots of the last token activations of layers 10 and 17 for Qwen2.5-0.5B-Instruct model. A geometric inversion can be observed in the Projections for the Uncertain and Certain input activations.}
    \label{fig:qwen-instruct-10and17layers}
\end{figure*}

 In all three models, we observe clear clustering patterns that validate the separability of epistemic modalities in the model’s internal representation space. Interestingly, in the later layers of the Instruct model (Layer 17 and 23), as well as in Layers 13 through 15 of the Chat model, we detect a geometric inversion in the PCA projections: the cluster corresponding to uncertain statements flips position relative to that of certain statements along the primary axes. {\it This inversion suggests a deeper semantic reorganization in the latent space, potentially signaling a transition from syntactic or lexical representation toward task-relevant abstractions of linguistic uncertainty.} These structured shifts reinforce our hypothesis that uncertainty is not only linearly encoded but is also semantically recontextualized in deeper layers, underscoring the interpretability and representational richness of the models under investigation.

\subsection*{How does Sensitivity to Linguistic Uncertainty change across layers?}
We extract activation vectors from all transformer layers using the TransformerLens library \citep{nanda2022transformerlens}, which allows access to cached internal states without modifying the model architecture. Specifically, for each input pair, consisting of a certain and an epistemically uncertain variant, we record the residual stream activations at the final token position, where model output is most strongly influenced.

To quantify the representational shift induced by linguistic uncertainty, we introduce a metric called \textit{Model Sensitivity to Uncertainty (MSU)}. This metric captures the average distance between the representations of the certain and uncertain variants of each input sentence pair.

Formally, for a given model layer $\ell$, we define MSU as:

\begin{equation}
\text{MSU}^{(\ell)} = \frac{1}{N} \sum_{i=1}^{N} \left\| \mathbf{h}_i^{(\ell, \text{certain})} - \mathbf{h}_i^{(\ell, \text{uncertain})} \right\|_2
\end{equation}

where $\mathbf{h}_i^{(\ell, \cdot)}$ denotes the activation vector obtained from layer $\ell$ for the $i$-th input in its certain or uncertain form, and $N$ is the total number of input pairs.

MSU provides a quantitative estimate of how much linguistic uncertainty perturbs the model's internal representations. Larger MSU values indicate greater sensitivity to uncertainty at that layer.

\subsection*{Results}

Across three models, we observe a strikingly consistent trend: the MSU scores increase monotonically with depth, indicating that sensitivity to epistemic uncertainty is a progressively emerging phenomenon in the transformer stack (Figure \ref{fig:msu}). Later layers exhibit substantially higher sensitivity to epistemic modals than early layers, indicating that semantic distinctions introduced by modality are progressively amplified across depth. This aligns with prior work indicating that deeper layers are responsible for encoding abstract, compositional semantics and final decision-making\citep{zhao-etal-2024-layer}. Our results suggest that epistemic uncertainty is treated as a high-level semantic feature.


\begin{figure}
    \centering
    \includegraphics[width=1\linewidth]{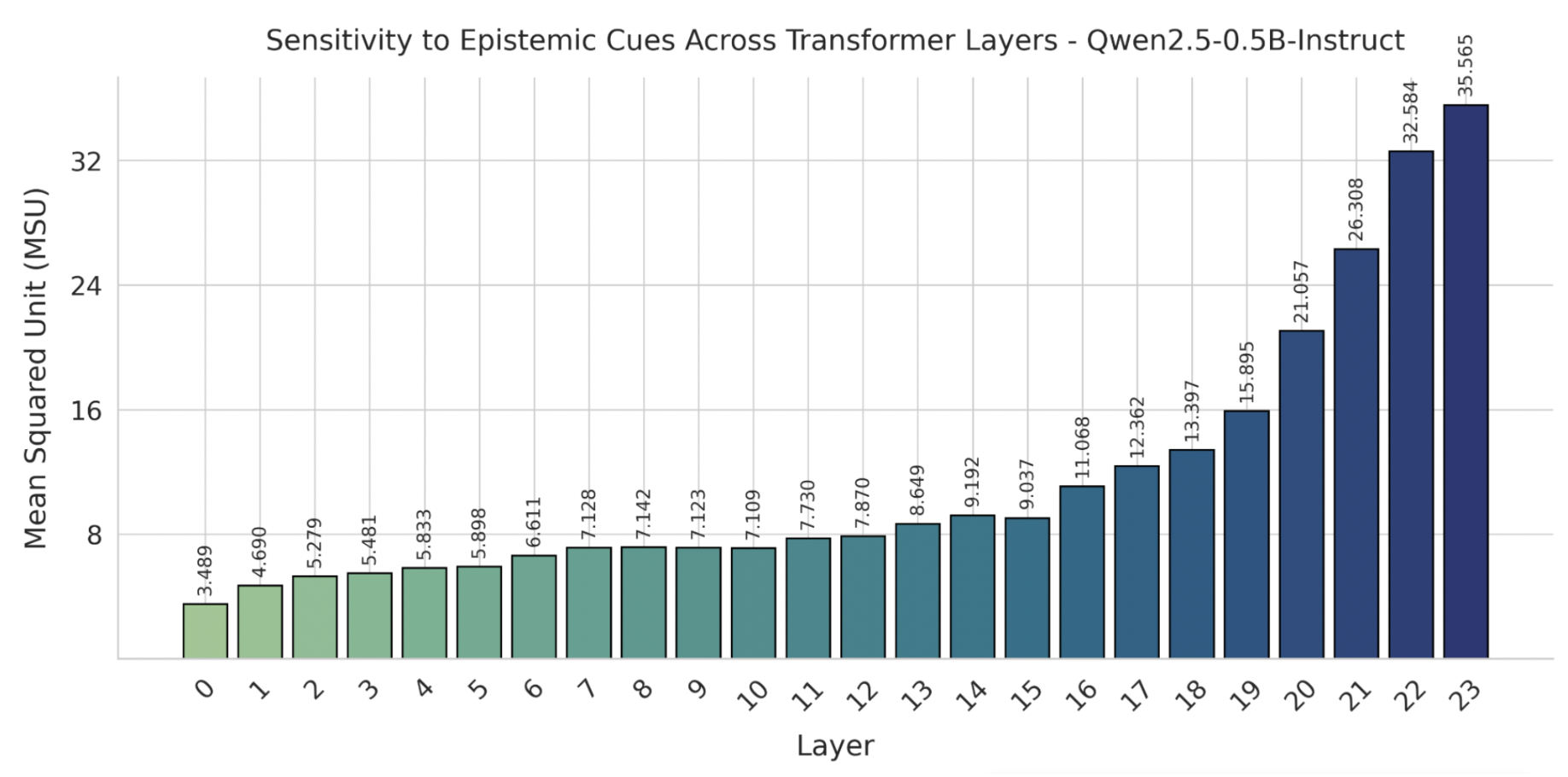}
    \caption{Layer-wise MSU scores for Qwen2.5-0.5B-Instruct, indicate progressively increasing scores across layers suggesting that later layers are responsible for encoding uncertainties in language}
    \label{fig:msu}
\end{figure}

\begin{table}[ht]
\centering
\begin{tabular}{|l|c|}
\hline
\textbf{Model} & \textbf{Average MSU} \\
\hline
LLaMA 3.2-1B & 9.361 \\
Qwen1.5-0.5B-Chat & \textbf{16.968} \\
Qwen2.5-0.5B-Instruct & 11.520 \\
\hline
\end{tabular}
\caption{Average MSU values across transformer layers for each model. Qwen1.5-0.5B-Chat comes out to be the most sensitive to linguistic uncertainty which is then followed by Qwen2.5-0.5B-Instruct and LLaMA 3.2-1B, among the 3 tested models.}
\label{tab:msu-averages}
\end{table}

Despite architectural and training differences between Qwen2.5 and Qwen1.5, both models demonstrate nearly identical MSU profiles, suggesting that this pattern of late-emerging sensitivity is robust to model family, scale, and instruction tuning. This implies that the encoding of linguistic uncertainty may be a universal behavior among autoregressive transformers

\section{Conclusion}
Our investigation reveals that the encoding of epistemic uncertainty is a distributed and emergent property of deep transformer architectures. Rather than residing in isolated layers, epistemic modality unfolds progressively, peaking in the final layers across models and variants alike. By proposing the Mean Sensitivity to Uncertainty (MSU) metric, we provide a targeted lens into this phenomenon. This structural consistency underscores a deeper semantic organization within LLMs and opens new pathways for designing models that are not only more interpretable, but also more epistemically aware.

\section*{Limitations}

This study takes an initial step toward understanding how language models encode linguistic uncertainty. Our findings are based on a limited set of instruction-tuned models, so their generality across architectures, sizes, and pretraining paradigms remains uncertain.

Future work should expand to include diverse model types, like non-instruct, multilingual, domain-specific and inputs namely, varied modal verbs, syntax, discourse). A key direction is localizing uncertainty to specific neurons or attention heads, and examining how internal uncertainty signals (e.g., logits, entropy) relate to output confidence and calibration.

\section{Acknowledgments}
We thank the developers of TransformerLens \cite{nanda2022transformerlens} for enabling detailed access to model internals and interpretability tools. We also acknowledge Anthropic for releasing the Persuasion dataset, which supported our exploratory evaluation. Additionally, we appreciate the Qwen team for open-sourcing their instruction-tuned models, which served as the basis for this study. This work was conducted independently without external funding.


\newpage
\bibliography{aaai2026}

\newpage

\appendix
\section{Appendix}
\label{sec:appendix}

\subsection{Layer-wise Sensitivity to Linguistic Uncertainty}
\label{sec:mcu-layerwise-allmodels}
Transformer-based language models adopt a deep, autoregressive architecture in which representations are refined layer by layer gradually building from lexical cues to nuanced semantic abstraction. This layered structure raises a key question: at which depth is linguistic uncertainty, especially that conveyed by epistemic modals (e.g., must, might, should, could), most saliently represented? To investigate this, we analyze the Mean Sensitivity to Uncertainty (MSU) across transformer layers for three Qwen2.5-0.5B variants Chat, Instruct, and Base as well as the LLaMA 3.2-1B-Instruct model.

Across all models, a consistent pattern emerges: early layers (e.g., Layers 0-5) demonstrate low sensitivity to uncertainty (MSU approx. 2.5-6), reflecting a focus on surface-level representations. In contrast, later layers (e.g., Layers 13-23) show a sharp rise in MSU, often surpassing 30, suggesting that deeper layers increasingly encode and amplify epistemic cues. For instance, in LLaMA 3.2-1B-Instruct, MSU climbs from 2.67 at Layer 0 to 31.96 at Layer 15. Among the Qwen variants, the Chat model displays the highest overall sensitivity, with elevated MSU values throughout the stack, indicating a heightened responsiveness to modal uncertainty. These findings support the hypothesis that the semantic abstraction required to capture epistemic nuance is a late-stage phenomenon within the transformer hierarchy.

\begin{figure}[H]
    \centering
\includegraphics[width=1.05\linewidth]{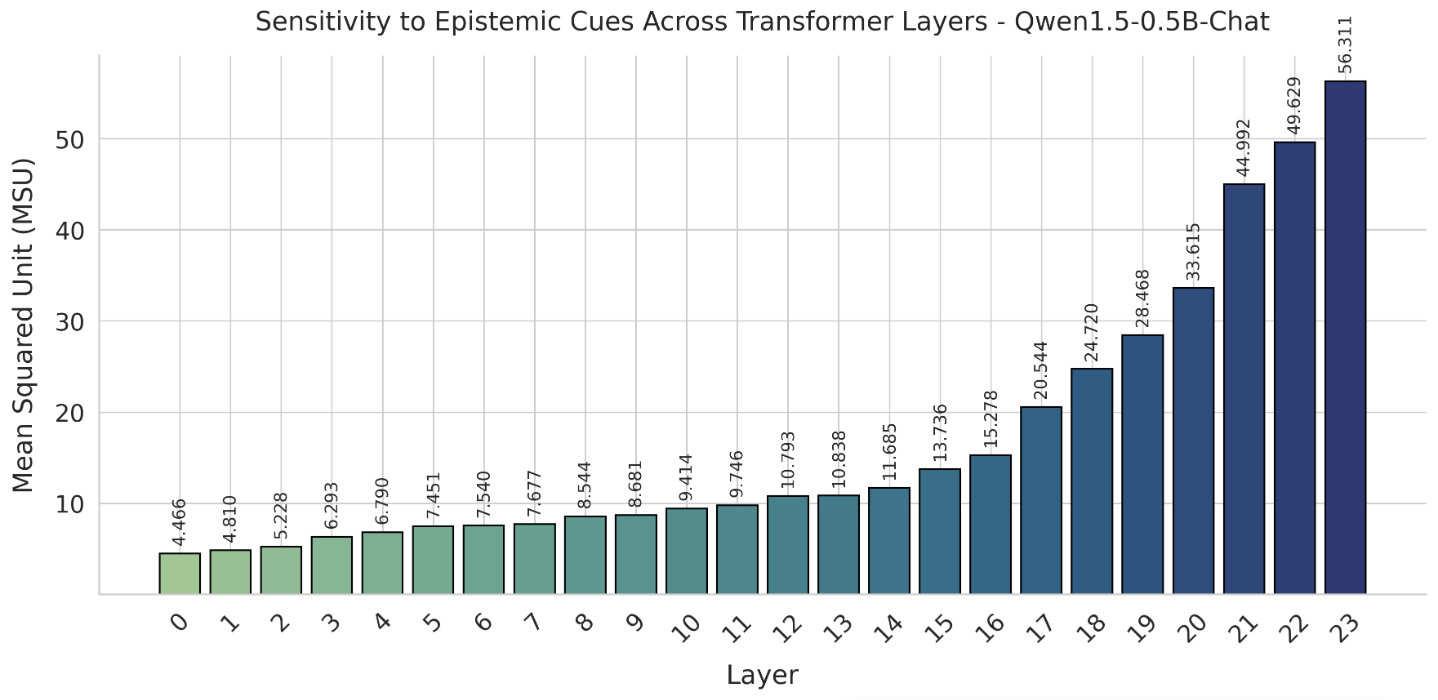}
    \caption{Layer-wise MSU scores for Qwen2.5-0.5B-Chat. Among all variants, the Chat model exhibits the highest sensitivity to linguistic uncertainty, with a steep increase in MSU across deeper layers.}
    \label{fig:msu}
\end{figure}

\subsection{PCA-analysis}
\label{sec: pca-analysis}
We observe a noticeable shift in the principal components of internal representations in the deeper layers of all models \ref{sec:model-specs} (Figures: \ref{fig:pca-layer-0}).
Specifically, while earlier and mid-level layers exhibit stable projection patterns, the final layers display a reorientation in the direction of PC1 and PC2. This structural transition likely reflects a late-stage reorganization of semantic or epistemic features, where uncertainty-related signals become more linearly separable or concentrated. Such emergent behavior may indicate that LLMs progressively consolidate abstract modality cues toward the final layers, where decision-critical information is encoded. This suggests that the geometry of representations not just their magnitude may carry functional signals related to epistemic reasoning.

\begin{figure}[H]
    \centering
    \includegraphics[width=1.05\linewidth]{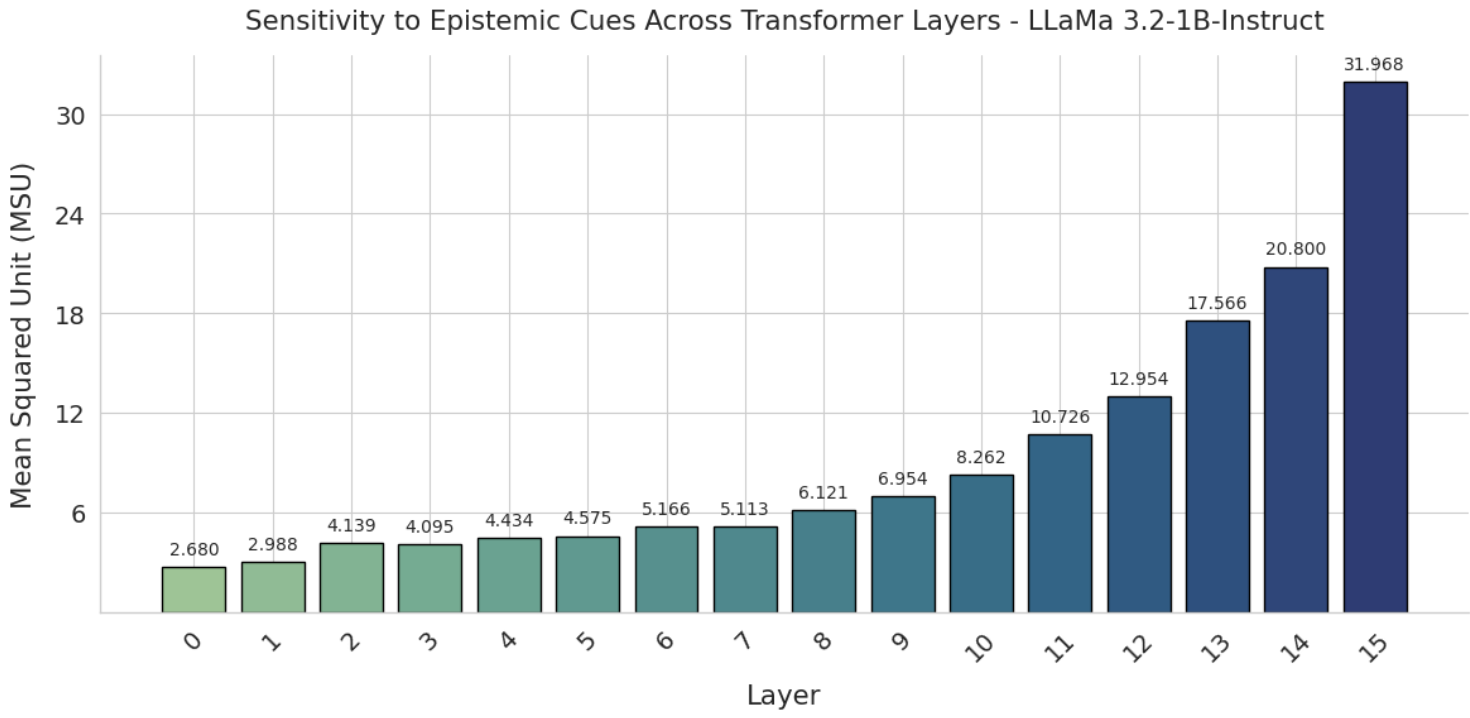}
    \caption{Layer-wise MSU scores for LLaMA 3.2-1B-Instruct, illustrating a steady increase in sensitivity to epistemic uncertainty across layers. The scores begin modestly in early layers (e.g., 2.68 at Layer 0) and rise sharply in deeper layers, peaking at 31.96 in Layer 15. This trend supports the hypothesis that later layers in autoregressive transformers are more attuned to modeling linguistic uncertainty.}
    \label{fig:enter-label}
\end{figure}

\subsection{Dataset Examples}
\begin{figure}[H]
    \centering
    \includegraphics[width=1\linewidth]{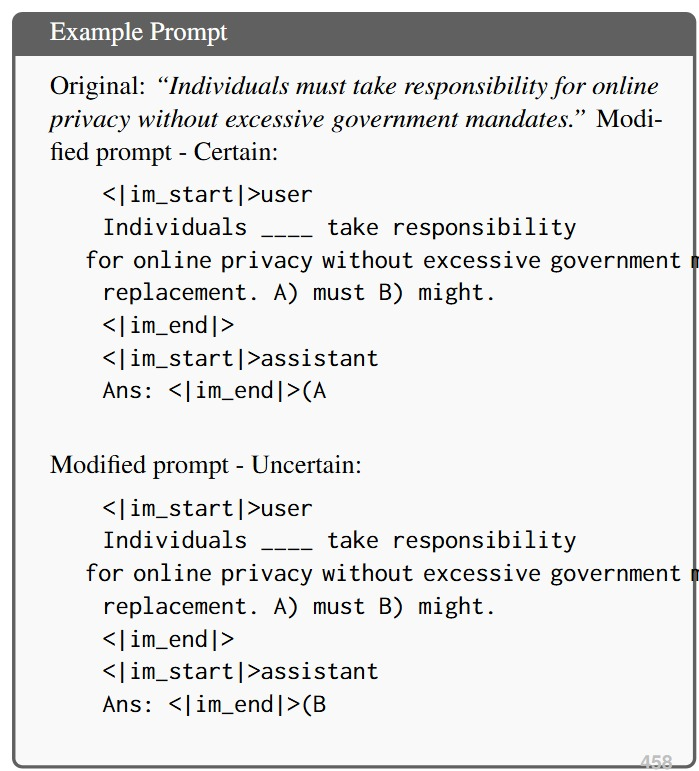}
    \caption{Example from dataset used to probe}
    \label{fig:dataset-eg1}
\end{figure}

\begin{figure}[H]
    \centering
    \includegraphics[width=1\linewidth]{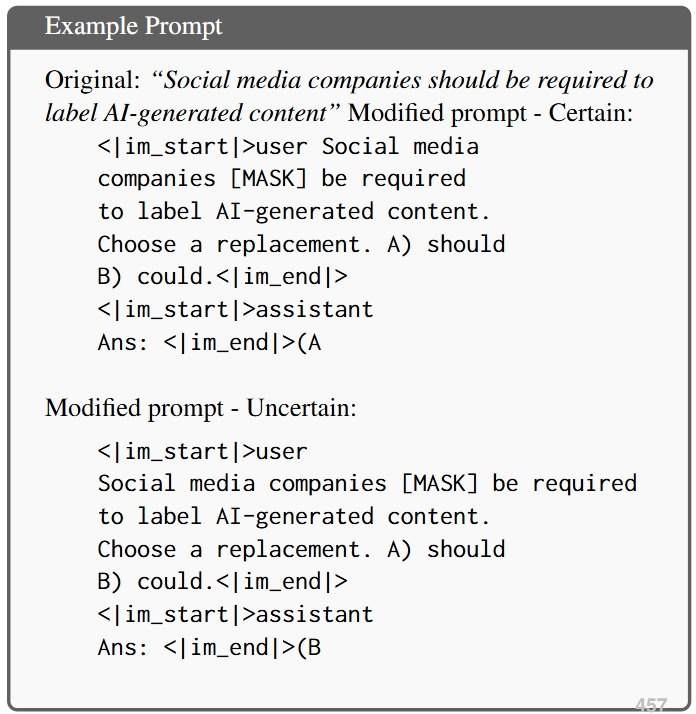}
    \caption{Example from dataset used to probe}
    \label{fig:dataset-eg1}
\end{figure}

\begin{figure}[t]
    \centering

    \begin{subfigure}{0.9\linewidth}
        \centering
        \includegraphics[width=\linewidth]{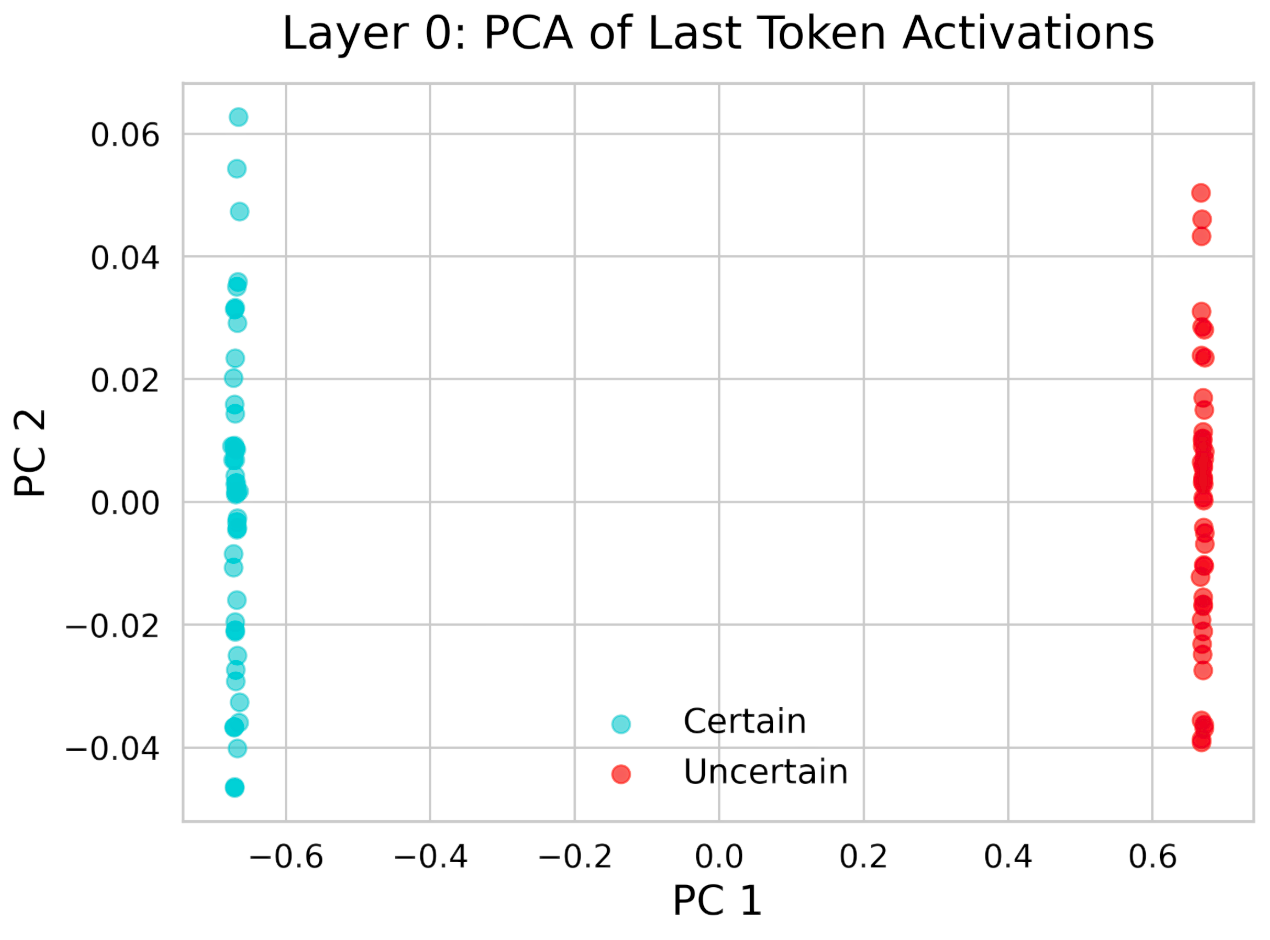}
        \caption{Layer 0}
        \label{fig:pca-layer-0}
    \end{subfigure}

    \vspace{1em}

    \begin{subfigure}{0.9\linewidth}
        \centering
        \includegraphics[width=\linewidth]{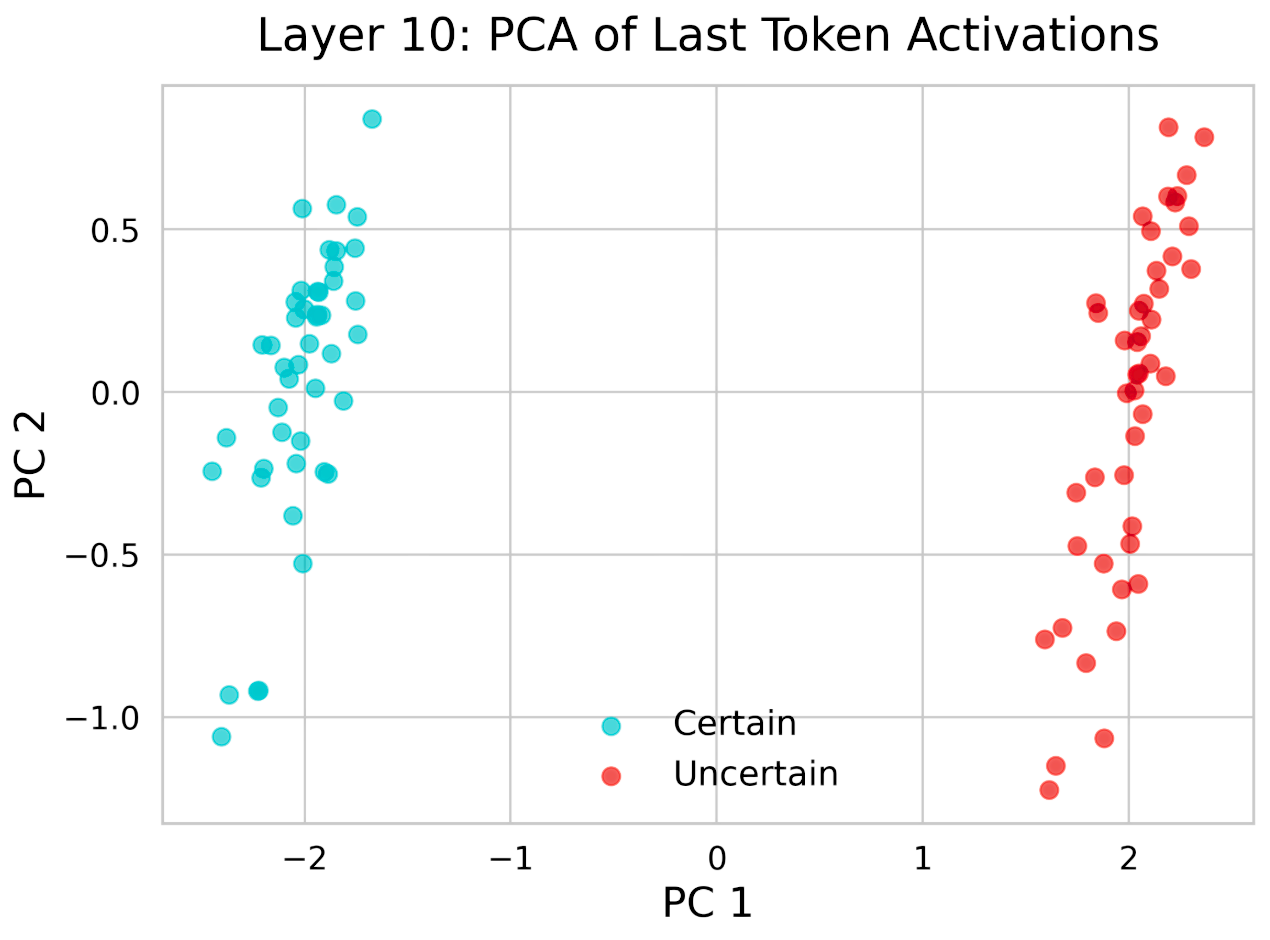}
        \caption{Layer 10}
        \label{fig:pca-layer-10}
    \end{subfigure}

    \vspace{1em}

    \begin{subfigure}{0.9\linewidth}
        \centering
        \includegraphics[width=\linewidth]
{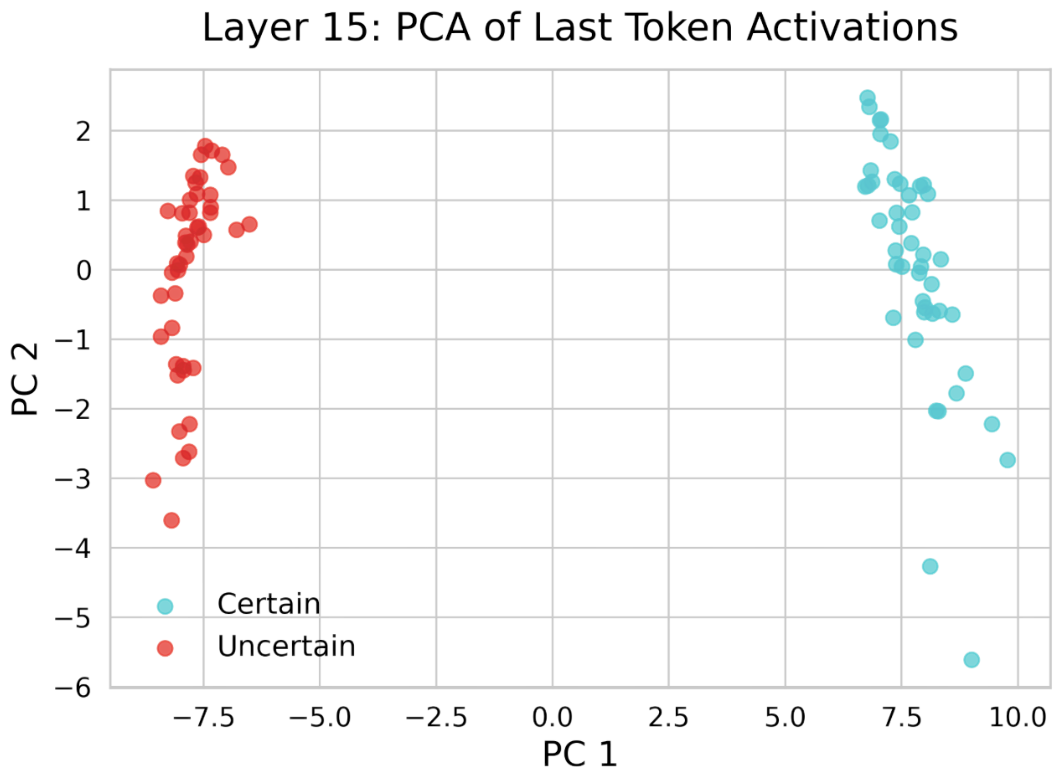}


        \caption{Layer 15}
        \label{fig:pca-layer-15}
    \end{subfigure}

    \caption{PCA projections of internal representations at different layers of the LLaMA 3.2 1B model. While early and mid layers exhibit relatively stable clustering patterns, the final layer shows a notable shift in the orientation of the principal components, suggesting a reorganization of the representational space. This directional change in PC1 vs.\ PC2 may reflect the model’s transition from encoding general contextual features to more task-specific or decision-relevant information.}
\end{figure}



\begin{figure*}[t]
\centering

\begin{subfigure}{0.40\linewidth}
    \centering
    \includegraphics[width=\linewidth]{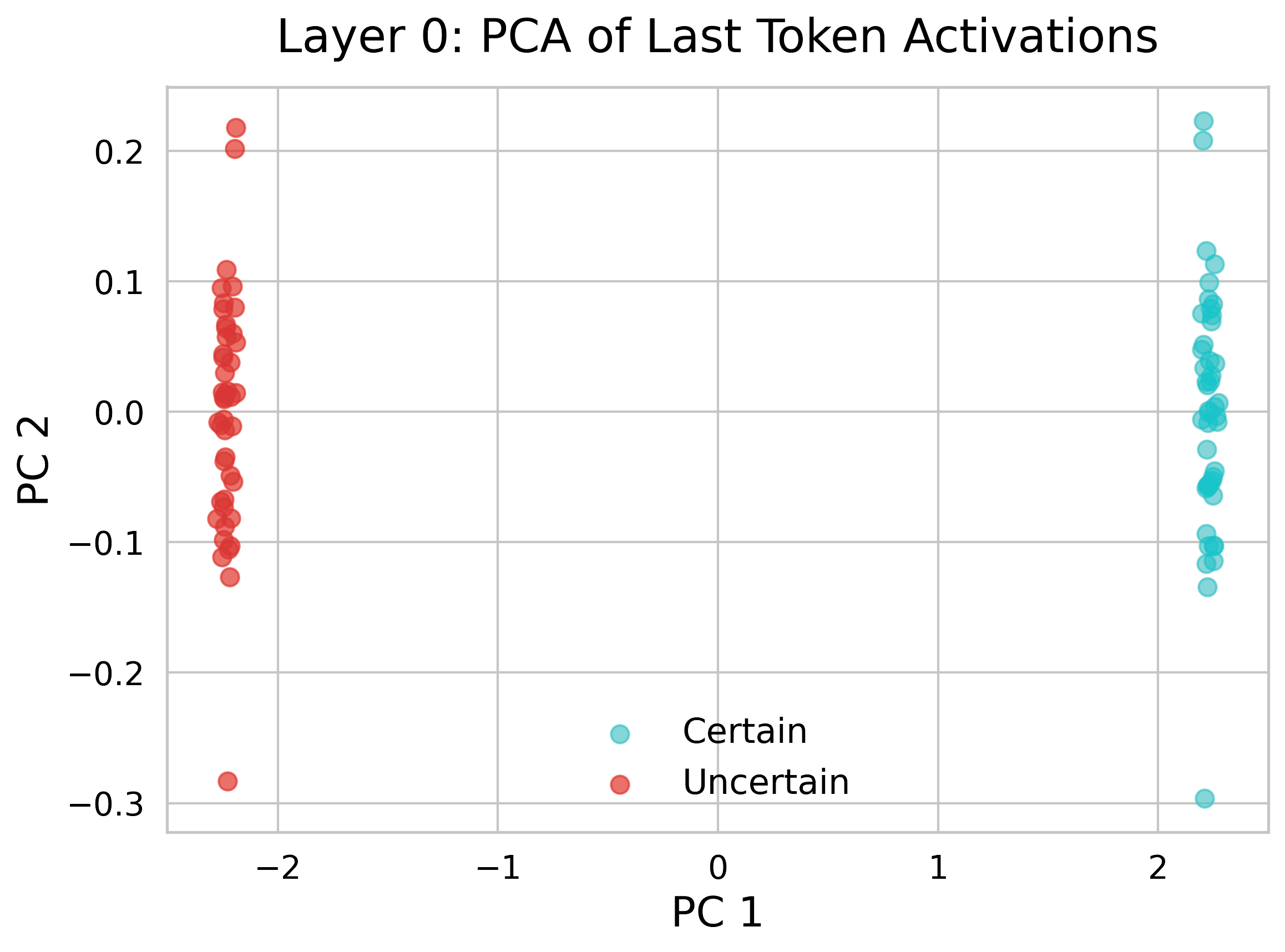}
    \caption{Layer 0}
    \label{fig:pca-layer-0}
\end{subfigure}
\hfill
\begin{subfigure}{0.40\linewidth}
    \centering
    \includegraphics[width=\linewidth]{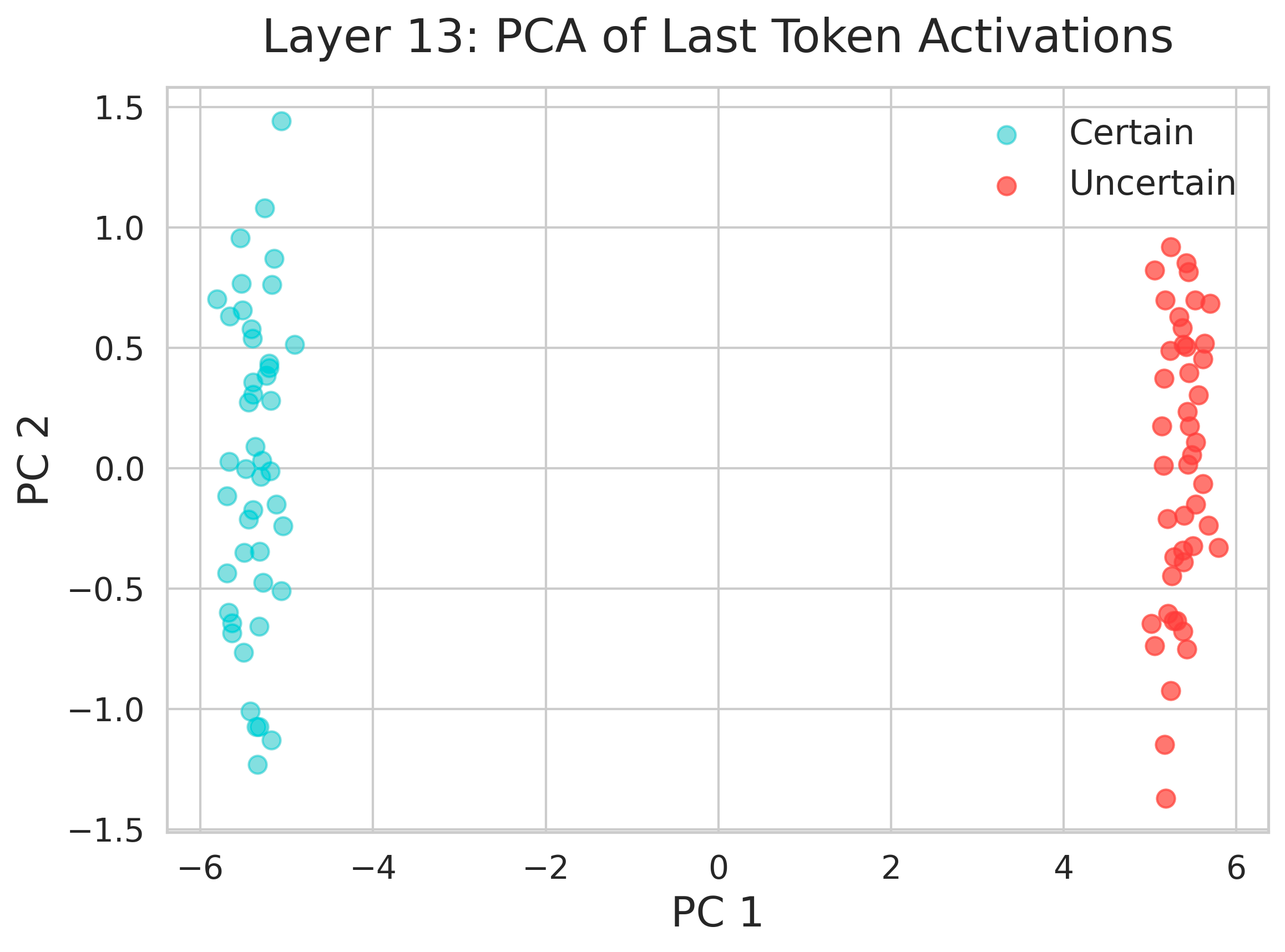}
    \caption{Layer 13}
    \label{fig:pca-layer-13}
\end{subfigure}

\vspace{1.5em}

\begin{subfigure}{0.40\linewidth}
    \centering
    \includegraphics[width=\linewidth]{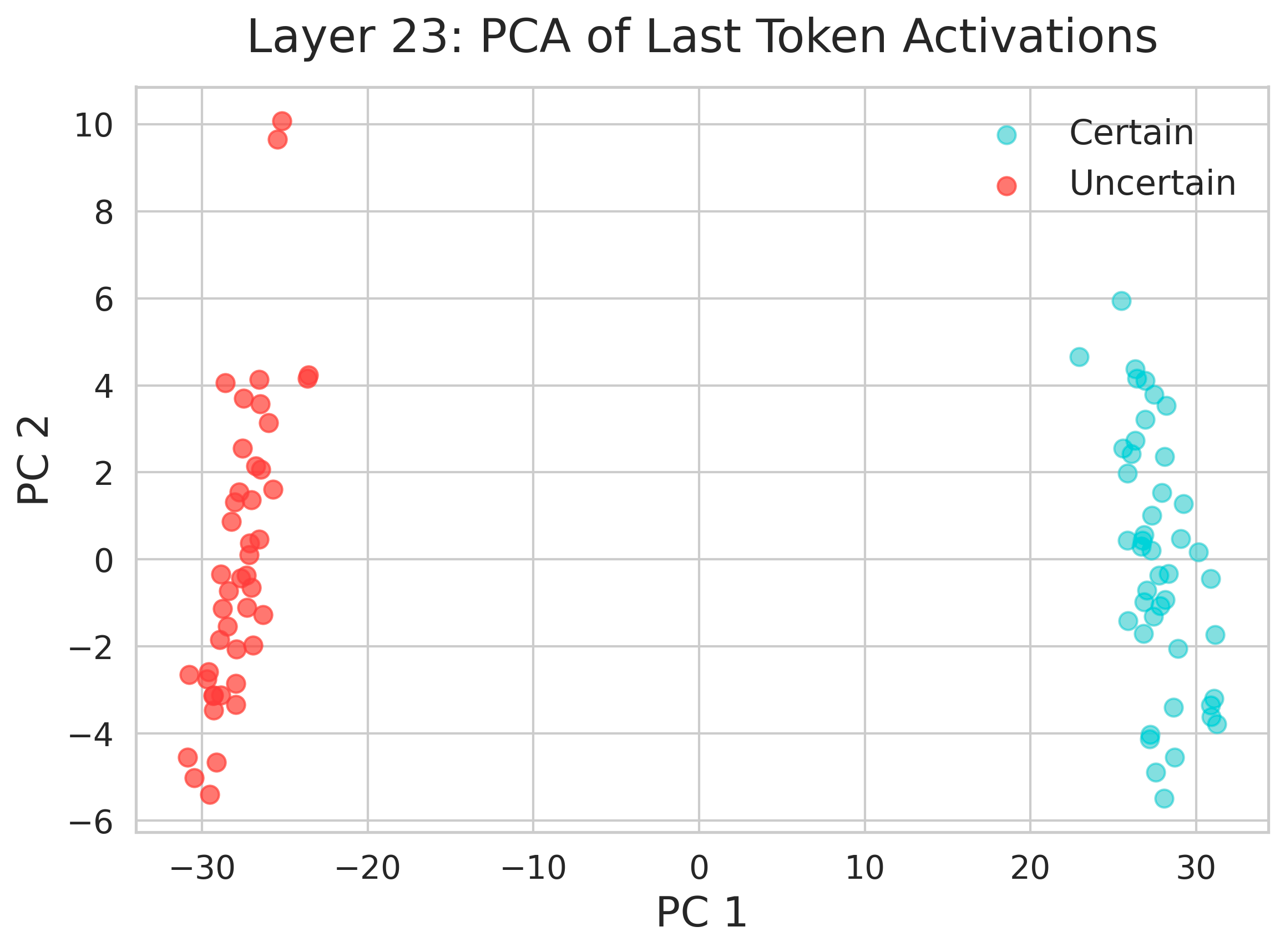}
    \caption{Layer 23}
    \label{fig:pca-layer-23}
\end{subfigure}
    \caption{PCA projections of internal representations at different layers of the Qwen1.5-0.5B-Chat model. Most layers display a consistent structure in the representational space; however, a marked shift in the direction of PC1 vs.\ PC2 emerges between layers 13 and 23. This transition suggests that epistemic information becomes reorganized or amplified in deeper layers, aligning with the model’s increasing sensitivity to uncertainty.}

    \label{fig:pca-all-layers}
\end{figure*}

\end{document}